\documentclass[final,iicol,natbib]{sn-jnl}

\usepackage{natbib}
\usepackage{algorithm}
\usepackage{algpseudocode}

\jyear{2022}%

\theoremstyle{thmstyleone}%
\theoremstyle{thmstyletwo}%

\theoremstyle{thmstylethree}%

\begin{document}

\title[Article Title]{Time series numerical association rule mining variants in smart agriculture}


\author*[1,2]{Iztok Fister Jr.}\email{iztok.fister1@um.si}

\author*[2]{Du\v{s}an Fister}\email{dusan@dusanfister.com}

\author[3]{Iztok Fister}\email{iztok.fister@um.si}

\author[4]{Vili Podgorelec}\email{vili.podgorelec@um.si}

\author[2]{Sancho Salcedo-Sanz}\email{sancho.salcedo@uah.es}

\affil*[1]{\orgdiv{Faculty of Electrical Engineering and Computer Science}, \orgname{University of Maribor}, \orgaddress{\street{Koro\v{s}ka cesta 46}, \city{Maribor}, \postcode{SI-2000}, \country{Slovenia}}}

\affil*[2]{\orgdiv{Department of Signal Processing and Communications}, \orgname{Universidad de Alcal\'a}, \city{Alcal\'a de Henares, Madrid}, \postcode{28805}, \country{Spain}}

\abstract{Numerical association rule mining offers a very efficient way of mining association rules, where algorithms can operate directly with categorical and numerical attributes. These methods are suitable for mining different transaction databases, where data are entered sequentially. However, little attention has been paid to the time series numerical association rule mining, which offers a new technique for extracting association rules from time series data. This paper presents a new algorithmic method for time series numerical association rule mining and its application in smart agriculture. We offer a concept of a hardware environment for monitoring plant parameters and a novel data mining method with practical experiments. The practical experiments showed the method's potential and opened the door for further extension.}




\maketitle

\section{Introduction}
Global food consumption is now at the highest level that it has ever been in history. Population growth (up to 9~billions till 2050 according to the FAO prediction~\cite{fao2009state}) and severe climate changes increase the need for food. Until recently, this problem was solved by increasing the role of crop production using mechanization, improved genetics, and increased inputs~\cite{collizi2020}. However, these increases have resulted mainly in depletion of soil, water scarcity, widespread deforestation and high levels of greenhouse gas emissions~\cite{fao2017soil,bajvzelj2014importance}. 

Despite the several negative reasons that impact the production of food, farmers are nowadays looking for a new and sustainable way for increasing food production. Smart farming is a paradigm for overtaking today's challenges to integrate two modern technologies, i.e., Information and Communication Technologies (ICT) \cite{sahitya2016wireless}, and the Internet of Things (IoT), in order to reorganize farming such that these disciplines and technologies could be involved in the smart way~\cite{collizi2020}. 
In the smart farming/agriculture vision, the land is equipped with different kinds of IoT sensors \cite{mohapatra2022ioe,agrawal2020improved}, capable of acquiring heterogeneous data. These data are transferred via sensor's rural networks to the Internet, where they are collected into complex databases, in which the knowledge necessary for analyzing the land characteristics is hidden. The intelligent algorithms, based on Artificial Intelligence (AI) \cite{issad2019comprehensive,dabre2018intelligent}, are employed for analyzing mined data in order to make rational decisions for the observed situations \cite{mishra2021modified,torres2021interpretability,fister2022time}. The decisions are transmitted either to the farmer's system in the form of actions or to the farmers in the form of messages \cite{ouafiq2022data}.

In this paper, we develop a hardware and software environment focused on computer vision for smart agriculture, where a Time Series Numerical Association Rule Mining (TS-NARM) algorithm is proposed and applied to tackle different problems arising in smart agriculture. The data are acquired from IoT sensors, which will measure different variables such as temperature, humidity, moisture, and light. On a time basis, these measurements are collected to time series frames that are mapped to features, with which the plant is monitored. Thus, each time series frame represents a transaction in a database. The transaction database serves as an origin for data analysis, in which time series data are identified and processed using TS-NARM. The algorithm mined a set of time base association rules that are ready to be explained to users by using the Explainable AI (XAI)~\cite{arrieta2020explainable}.

The purpose of the study is therefore twofold: (1) to develop the data collection and preprocessing method, and (2) to propose a TS-NARM based system to process the data and knowledge extraction. This paper is distinguished by the following main novel contributions:
\begin{itemize}
    \item a lightweight method for data acquisition based on an ESP32 micro-controller is developed, which includes several sensors for capturing significant data and environmental variables.
    \item a comprehensive collected dataset has been obtained, which allows a further treatment of the data via AI techniques.
    \item stochastic nature-inspired algorithms for TS-NARM construction are developed, while a comprehensive comparative study is performed, in order to show their advantages and shortcomings.
\end{itemize}

The structure of the remainder of the paper is as follows: Section~\ref{sec:2} is dedicated to explain the background information necessary to potential readers for understanding the topics that follow, including concepts on association rules mining and evolutionary algorithms. In Section~\ref{sec:3}, the experimental setup is illustrated, where the concept of the proposed smart agriculture is introduced, together with the laboratory setup of the hardware, as well as the developed algorithms for TS-NARM. The results of the experiments are the subject of Section~\ref{sec:4}. The paper concludes with a discussion in Section~\ref{sec:5}, which summarizes the performed work and outlines directions for the future. 

\section{Background information}\label{sec:2}

\subsection{Association rule mining}
This section briefly presents the formal definition of ARM. Let us suppose a set of objects $O=\{o_1, \ldots,o_M\}$, where $M$ denotes the number of attributes, and transaction set $D$ are given, where each transaction $Tr$ is a subset of objects, in other words $Tr \subseteq O$. Then, an association rule can be defined as implication:
\begin{equation}
X \implies Y, 
\label{arule}
\end{equation}
\noindent where $X \subset O$, $Y \subset O$, in $X \cap Y = \emptyset$. The following two measures are defined for evaluating the quality of the association rule~\cite{agrawal1994fast}:
\begin{equation}
\mathit{conf}(X \implies Y) = \frac{n(X \cap Y)}{n(X)}, 
\label{Eq:1a}
\end{equation}
\begin{equation}
\mathit{supp}(X \implies Y) = \frac{n(X \cap Y)}{N}, 
\label{Eq:2a}
\end{equation}
\noindent where $\mathit{conf}(X \implies Y)\geq C_{\mathit{min}}$ denotes confidence and $\mathit{supp}(X \implies Y) \geq S_{\mathit{min}}$ support of the association rule $X \implies Y$. Thus, $N$ in Eq.~(\ref{Eq:2a}) represent the number of transactions in transaction database $D$ and $n(.)$ is the number of repetitions of a particular rule $X \implies Y$ within $D$. Here, $C_{\mathit{min}}$ denotes minimum confidence and $S_{\mathit{min}}$ minimum support. This means that only those association rules with confidence and support higher than $C_{\mathit{min}}$ and $S_{\mathit{min}}$ are taken into consideration, respectively.

In order to control the quality of the mined association rules in more detail, two additional measures are defined, i.e., inclusion and amplitude. Inclusion is defined as the ratio between the number of attributes of the rule and all the attributes in the database~\cite{hasler2007new}:
\begin{equation}
    \textit{incl}(X \implies Y) = \frac{|X|+|Y|}{M},
\end{equation}
where $M$ is the total number of attributes in the transaction database. Amplitude measures the quality of a rule, preferring attributes with smaller intervals, in other words~\cite{fister2021improved}:
\begin{equation}
\small
    \textit{ampl}(X \Rightarrow Y) = 1 - \frac{1}{M}\sum_{k = 1}^{m}{\frac{\textit{Ub}_k - \textit{Lb}_k}{\max(o_k) - \min(o_k)}},
\end{equation}
where $Ub_k$ and $Lb_k$ are the upper and lower bounds of the selected attribute, and $\max(o_k)$ and $\min(o_k)$ are the maximum and minimum feasible values of the attribute $o_k$ in the transaction database.

\subsection{Stochastic population-based nature-inspired algorithms}
Stochastic population-based nature-inspired algorithms are a common name comprising two families of optimization algorithms under the same umbrella, i.e., Evolutionary Algorithms (EAs) and Swarm Intelligence (SI) based algorithms. The characteristics of these are already hidden in their name. This means that they are stochastic in nature, due to employing a random generator by constructing new, potentially better solutions. In place of searching for a single solution, they explore the knowledge hidden within the whole population of solutions. The final characteristic, i.e., nature-inspired, refers to an inspiration taken from nature, on which their search process is found~\cite{del2019bio,tzanetos2021nature}.

In our study, both kinds of algorithms are applied for solving the TS-ARM in smart agriculture. Therefore, the similarity and differences of both families are discussed in a nutshell in the remainder of the paper.

\subsubsection{Evolutionary algorithms}
EAs are metaheuristic approaches based on the evolution of natural species \cite{del2019bio}. According to this theory, the fitter individuals have more chances to survive in unpleasant environmental conditions due to their better adaptation to them. Thus, the less fit ones are eliminated by the natural selection. Indeed, all individuals' characteristics are written in their genes (i.e., genotype) that are inherited from generation to generation, while their traits (i.e., phenotype) are reflected from the genotype. The genetic material is transferred to the next generations via a process of reproduction consisting of crossover and mutation~\cite{eiben2015introduction}. In this way, the crossover serves for mixing the genetic material between parents, while the mutation takes care of the diversity of the material.

The evolutionary process has became an inspiration for developing the EAs. Similar to natural processes, EAs also consist of populations of individuals representing solutions of the problem to be solved. The natural population suffers under conditions of dynamical environment changing constantly over time. This environment is presented in EAs by the problem, to which optimal solutions are drawn nearer by exploring the problem's search space. Thus, the offspring solutions undergo the effects of acting the crossover and mutation operators. Finally, the quality of each individual is estimated using the evaluation function.

Algorithm~\ref{alg:eas} illustrates a pseudo-code of the common EAs. As can be seen from the pseudo-code,
\begin{algorithm}[htb]
\caption{Evolutionary algorithm}
\label{alg:eas}
\begin{algorithmic}[1]
\State INITIALIZE\_population\_randomly
\State EVALUATE\_each\_individual
\While {TERMINATION\_CONDITION\_not\_met} 
\State SELECT\_PARENTS
\State CROSSOVER\_parents
\State MUTATE\_offspring
\State EVALUATE\_each\_offspring
\State SELECT\_SURVIVALS
\EndWhile
\end{algorithmic}
\normalsize
\end{algorithm}

An evolutionary cycle starts with an initialization of a population of solutions, normally, represented as binary, integer, or real-valued vectors (line~1). After initialization, the evaluation of solutions is launched (line~2). Then, the \textbf{while} loop introduces the evolutionary cycle (lines~3-9). that is terminated with the termination condition. In each evolutionary cycle, the parent selection operator selects two parents, which contribute to mixing their genetic material with the crossover and mutation operators by creating new offspring (lines~5-6). Next, the quality of offspring is evaluated with the fitness function (line~7). Finally, the survival selection operator determines those members of the current population that will transfer their genetic material to the next generations. 

Moreover, the family of EA-based approaches is large, and consist of many different approaches \cite{del2019bio}, among others:
\begin{itemize}
\item Genetic Algorithms (GA)~\cite{goldberg2013genetic},
\item Genetic Programming (GP)~\cite{koza1992genetic},
\item Evolution Strategies (ES)~\cite{rechenberg1973ingo},
\item Evolutionary Programming (EP)~\cite{fogel1966artificial},
\item Differential Evolution (DE)~\cite{storn1997differential}.
\end{itemize}
Although all the aforementioned algorithms follow the common principle of EAs as illustrated in Algorithm~\ref{alg:eas}, they differ between each other regarding the representation of individuals. For instance, the individuals in GAs are represented as binary strings, while, in the GP, as programs in the Lisp programming language. The final state automata form a population of solutions in EP, while the real-valued vectors appear in the role of population members in ES and DE.

\subsubsection{Swarm intelligence-based algorithms}
Inspiration for SI-based algorithms has also been drawn from the nature, precisely, from collective behavior in biological systems~\cite{blum2008swarm}. For, instance, some kinds of insects (e.g., honeybees and ants) and animals (e.g., fishes and birds) live in a society, e.g., honeybee's combs, ant colonies, schools of fish, and flocks of birds. Thus, they expose the swarm intelligence in the following sense: Although the particles (also agents) of swarms are capable of performing only simple tasks, they can deal with complex problems together as a group. In line with this, decision-making in a swarm is decentralized, while the particles are capable of self-organization. They interact between each other using some kind of communication that can be either direct or indirect~\cite{fister2015adaptation}. In the former case, information is transmitted without the intervention of the environment, while, in the latter case, individuals are not in direct contact, because the communication is conducted via environmental data.

Similar as in EAs, the SI-based algorithms also operate with a population of solutions that is called a swarm of particles in the sense of SI. The particles represent solutions of the problem to be solved, and are, typically, defined as real-valued vectors~\cite{fister2022dynfs}. During the optimization cycle, they move within the problem search space towards the better ones, and, in this way, discover new, potentially better solutions. Normally, the moves are described regarding the physical equations that mimic the moves of particles in natural biological systems. Also here, only the best particles are selected for the next generations, while the optimization cycle is terminated using a termination condition.

The pseudo-code of the SI-based algorithms is illustrated in Algorithm~\ref{alg:SI}~\cite{engelbrecht2005fundamentals}, from which it can be seen  
\begin{algorithm}[htb]
\caption{Swarm intelligence}
\label{alg:SI}
\small
\begin{algorithmic}[1]
\State INITIALIZE\_population\_randomly
\State EVALUATE\_each\_particle
\While {TERMINATION\_CONDITION\_not\_met} 
\State MOVE\_towards\_better\_particle
\State EVALUATE\_each\_particle
\State SELECT\_SURVIVALS
\EndWhile
\end{algorithmic}
\normalsize
\end{algorithm}
that it differs from Algorithm~\ref{alg:eas} in line~4, where the move operator is applied in place of parent selection and variation operators as in EAs (lines 4-6).

Until end of the last decade, a flood of newly developed SI-based algorithms has been emerging that raised criticism in the nature-inspired community~\cite{sorensen2015metaheuristics} about the question how novel these algorithms were and if they did not hide behind their famous metaphor taken from nature's inspiration. The critics slowed down the flood, and, nowadays, only the more valuable algorithms can find a way to the research community. Although the majority of the SI-based algorithms are represented with real-valued vectors~\cite{fister2022dynfs} and, therefore, the classification to this criteria, as by EAs, is not possible, one of the first tries to classify them was proposed in~\cite{fister2013brief}. Actually, this classification was based on their inspirations from nature.   

\subsection{NiaPy framework}
A NiaPy library~\cite{vrbancic2018niapy} is a framework of nature-inspired algorithms implemented in Python programming language. This package is distributed under the MIT licence, and enable potential developers to avoid the implementation of these algorithms, which can sometimes be a difficult, complex, and tedious task. The implementations of algorithms in the library are verified, while their codes comply with the last Python standards. Currently, the library consists of 29 original nature-inspired, 7 modified, and 6 other algorithms.

Together with the aforementioned algorithms, a lot of test problems are also appended into the library. This fact enables the users to compare various algorithms between each other easily, and helps them to decide which algorithm to apply for solving their practical problems. Due to its simplicity of use, this library has also become an unavoidable tool for comparing the different nature-inspired algorithms at various universities around the world.

\section{Experimental environment}\label{sec:3}
In this section, we present our experimental environment, that involved a hardware unit consisting of three sensors, which allowed us to acquire data, all software and hardware components used for data collection, and the data preprocessing techniques applied to them.

\begin{figure*}[htb]
\centering
\includegraphics[width=10cm]{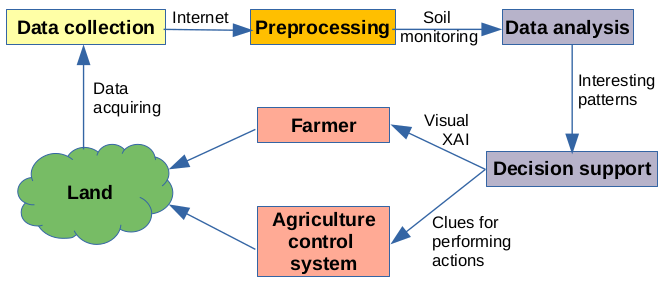}
\caption{Concept of the smart agriculture.}
\label{hardware1}
\end{figure*}

The concept of the smart agriculture in our study is illustrated in Fig.~\ref{hardware1}, from which it can be seen
that different IoT sensors monitor the land characteristics. Via a rural network, they are connected to a network access point, that serves for data collection and enables them access to the Internet. The collected data are reduced and preprocessed, in order to map  only those indicators to extracted features that refer to soil monitoring. Obviously, each data entry is supplemented with its date and time information. Such data then enter into data analysis, in which interesting patterns (also knowledge) are mined. The decision-making process is started based on the interesting patters. The results of this process can be represented in two ways: (1) to explain unexplained data, and (2) to propose clues for performing actions. The former serve as an input to the XAI that suggests to the farmer what to do in a specific situation, while the latter proposes an action that could to be performed by the agriculture controlled system (e.g., start to irrigate a plant for 10 minutes). Let us notice that the study is focused only on the data collection, preprocessing, and data analysis. Due to the complexity of XAI, the last step remains a subject of the future work. 

Implementing the concept of smart agriculture demands hardware and software components that must be integrated into a control system. In summary, the system in smart agriculture consist of the following components: 
\begin{itemize}
    \item hardware unit,
    \item data collection,
    \item data preprocessing,
    \item TS-ARM with nature-inspired algorithms.
\end{itemize}
In the remainder of the paper, the aforementioned components are illustrated in detail.

\subsection{Hardware unit}
\begin{figure*}[htb]
    \centering
    \includegraphics[width=0.75\textwidth]{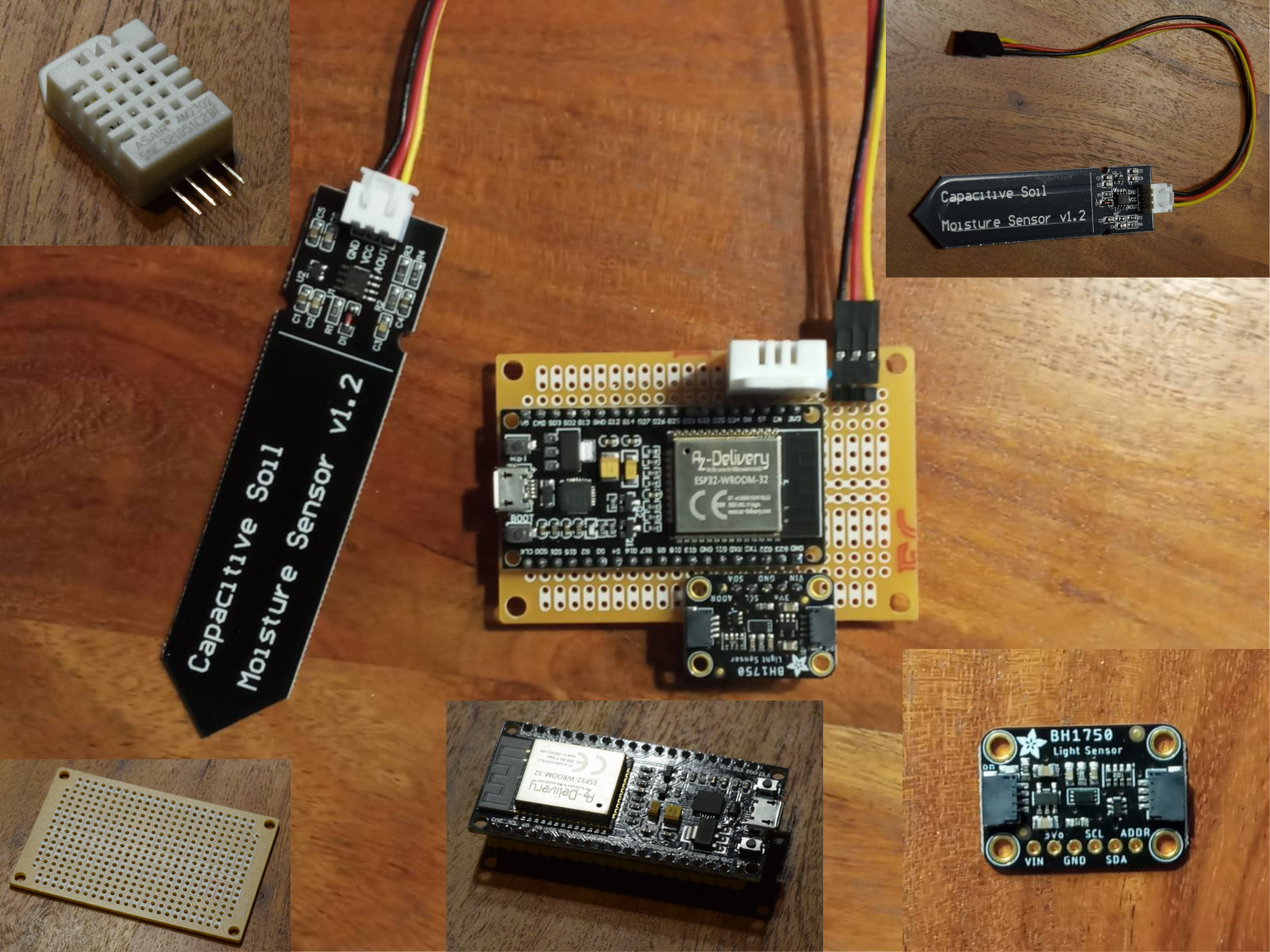}
    \caption{A sketch photo of used elements.}
    \label{fig:sketch}
\end{figure*}

The hardware unit consists of sensors connected into a rural network, and an access point for acquiring data from the sensors and transmitting them to the Internet. Thus, the prototype hardware unit was built. Table~\ref{tab:hw} lists all the hardware components that were used in our solution. 
\begin{table*}[htb]
\centering
\caption{Specification of hardware equipment.}
\label{tab:hw}
\centering
\begin{tabular}{ c|l|l } 
Ind. & Component & Function \\
 \hline
 1 & ESP32 NodeMCU Module & microcontroller \\
 2 & Adafruit BH1750 & light intensity sensing \\ 
 3 & DHT22 AM2302 & air temperature and humidity sensing\\
 4 & Soil Moisture Hygrometer Module & soil moisture sensing\\
 \hline
\end{tabular}
\end{table*}
All the applicable sensors have been welded permanently to a simple perfboard for the sake of proof-of-concept, and wired to the ESP32 NodeMCU module. Standard communication protocols were utilized. Figure~\ref{fig:sketch} visualizes a collage of the individual elements.

Actually, the ESP32 module represents the heart of the system and enables processing power for the data collection. The data are obtained via an Adafruit BH1750 light intensity sensor, DHT22 AM2302 air temperature and humidity sensor, and Soil Moisture Hygrometer sensors. Data are transferred to the webserver in predefined time periods, where the data are stored in a database.

\subsection{Data collection}
Data from the sensors, also Sensor Data (SD), are acquired as a tuple:
\begin{equation}
\begin{aligned}
    \mathit{SD}= & \langle \mathit{Light},\mathit{Temperature},\mathit{Humidity}, \\
    & \mathit{Moisture},\mathit{Date},\mathit{Time} \rangle,
\end{aligned}
\end{equation}
where the light, temperature, humidity, and moisture indicators are obtained from the corresponding sensors. 

Actually, the tuples $SD$ are acquired in a specific time period that are defined by the user. Thus, it holds, the shorter the time period, the more detailed acquired data. These are transmitted to the Internet server using a straightforward Python application running on the web server, pprocessing the HTTP requests utilizing a web.py library.

\subsection{Data preprocessing}
Data preprocessing is usually one of the most critical steps in the whole data science process. Data preprocessing can be defined as a set of methods that enhance the overall quality of the raw data and try to enrich it~\cite{fan2021review,fister2022dynfs,fister2022time}. Essentially, two tasks are required in time series data preprocessing phase: 
\begin{itemize}
\item data reduction,
\item feature extraction.
\end{itemize}
The first preprocessing task enables grouping the data in time frames, while the second is devoted to data enrichment.

Time series $TS$ is defined as a sequence of the collected data tuples $SD_i$ for $i=1,\ldots,T$:
\begin{equation}
    TS=SD_1,\ldots,SD_T,
\end{equation}
where $T$ denotes the number of data tuples in time series (also time series size).  

The lack of measured indicators prevents the TS-NARM to produce any specific insights. Therefore, we must enrich collected data by additional features reflecting a better outlook on time-series data. 
Time series Frame $TF$ is obtained by a data reduction ML preprocessing method, where it is expected that the method analyzing $TF$ provides the same results as analyzing the original $TS$. In line with this, a set of indicators collected in $TS$:
\begin{equation}
\small
\begin{aligned}
    \mathit{INDICATOR}=&\{\mathit{LIGHT},\mathit{TEMPERATURE},\\
    & \mathit{HUMIDITY},\mathit{MOISTURE},\\
    & \mathit{DATE},\mathit{TIME}\},
\end{aligned}
\end{equation}
is reduced by a set of modifiers:
\begin{equation}
    \mathit{MODIFIER}=\{\mathit{MIN,MAX,AVG,DIF}\}.
\end{equation}
In order to determine a set of compound features, a Cartesian product of sets $MODIFIER$ and $INDICATOR$ is calculated except for the indicators $TIME$ and $DATE$. The results of the feature extraction is illustrated in Table~\ref{tab:cat1},
\begin{table*} 
\caption{Defined features.} 
\label{tab:cat1}
\begin{tabular}{|c|l|l|l|}
\hline
Nr. & Feature & Attribute domain & Short description \\ \hline
1 & AVG\_TEMPERATURE & NUMERIC & Average temperature of data in the TF \\ 
2 & MAX\_TEMPERATURE & NUMERIC & Maximum temperature of data in the TF \\ 
3 & MIN\_TEMPERATURE & NUMERIC & Minimum temperature of data in the TF \\ 
4 & DIF\_TEMPERATURE & NUMERIC & Temperature interval of data in the TF \\ \hline
5 & AVG\_HUMIDITY & NUMERIC & Average humidity of data in the TF \\ 
6 & MAX\_HUMIDITY & NUMERIC & Maximum humidity of data in the TF \\ 
7 & MIN\_HUMIDITY & NUMERIC & Minimum humidity of data in the TF \\ 
8 & DIF\_HUMIDITY & NUMERIC & Humidity interval of data in the TF \\ \hline
9 & AVG\_MOISTURE & NUMERIC & Average moisture of data in the TF \\ 
10 & MAX\_MOISTURE & NUMERIC & Maximum moisture of data in the TF \\ 
11 & MIN\_MOISTURE & NUMERIC & Minimum moisture of data in the TF \\ 
12 & DIF\_MOISTURE & NUMERIC & Moisture interval of data in the TF \\ \hline
13 & AVG\_LIGHT & NUMERIC & Average light of data in the TF \\ 
14 & MAX\_LIGHT & NUMERIC & Maximum light of data in the TF \\ 
15 & MIN\_LIGHT & NUMERIC & Minimum light of data in the TF \\ 
16 & DIF\_LIGHT & NUMERIC & Light interval of data in the TF \\ \hline
17 & SEQUENCE & NUMERIC & Time series sequence  \\ 
18 & CLASS & NUMERIC & Class of the time series \\ 
\hline
\end{tabular}
\end{table*}
where each compound feature is represented as a concatenation of $MODIFIER\times INDICATOR$ denoted by a character '\_', while indicator $DATE$ is mapped to the feature $SEQUENCE$ and the indicator $TIME$ to the feature $CLASS$. Thus, the modifiers are mathematically defined as follows:
\begin{equation}
\footnotesize
\begin{aligned}
    \mathit{MIN\_INDICATOR}&=\min_{i=1,\ldots,T} \mathit{SD_i.INDICATOR},\\
    \mathit{MAX\_INDICATOR}&=\max_{i=1,\ldots,T} \mathit{SD_i.INDICATOR},\\
    \mathit{AVG\_INDICATOR}&=\frac{1}{T}\sum_{i=1}^T \mathit{SD_i.INDICATOR},\\
    \mathit{DIF\_INDICATOR}&=\frac{1}{2}(\mathit{SD_T.INDICATOR}+\\
    &\mathit{SD_1.INDICATOR}),
\end{aligned}
\end{equation}
where $SD_i.INDICATOR$ for $i=1,\ldots,T$ specifies particular indicator collected by $i$-th frame of the specific TS. While the definition of the first three modifiers is self-explanatory, the modifier $DIF\_INDICATOR$ is expressed as an difference of the indicator measured at the end and the beginning the time period and thus highlights a variance of the values within the TS. The feature $SEQUENCE$ is calculated such that the starting date is attached to value $SEQUENCE=0$, and then the value is incremented by one for each next date. The indicator $TIME$ in the form $hh:mm:ss$ is mapped firstly to a timestamp $timestamp$ as:
\begin{equation}
    timestamp=hh*3600+mm*60+ss,
\end{equation}
and then to the proper feature $CLASS$ according to the following equation:
\begin{equation}
    CLASS=\left\lfloor\frac{timestamp}{86400}\cdot K\right\rfloor+1,
\end{equation}
where $K$ denotes the number of time intervals, into which the 24-hour period (i.e., 86,400~sec) is divided. The selection of the proper value of $K$ is crucial for the results of the optimization.

In summary, the time series database $D$ of dimension $N\times M$, where $N$ denotes the number of transactions in the database, and $M$ is the number of features, where each transaction is defined as a sequence of the features defined in Table~\ref{tab:cat1}. 

\subsection{Time Series Association Rule Mining with nature-inspired algorithms}
The purpose of this section is to present the mathematical foundations of TS-ARM and the necessary modifications that must be applied to nature-inspired algorithms for implementing TS-ARM. In our study, the following nature-inspired algorithms are applied:
\begin{itemize}
    \item Differential Evolution (DE)~\cite{storn1997differential},
    \item Genetic Algorithm (GA)~\cite{goldberg2013genetic},
    \item Particle Swarm Optimization (PSO)~\cite{kennedy1995particle},
    \item Success-history based adaptive differential evolution using linear population size reduction (LSHADE)~\cite{tanabe2014improving},
    \item self-adaptive differential evolution (jDE)~\cite{brest2006self}.
\end{itemize}
Actually, two components of nature-inspired algorithms need to be modified by implementation of the TS-ARM, i.e., representation of solutions and fitness function. Let us mention that the implementations of the original aforementioned algorithms are taken from NiaPy library.

\subsection{Time Series Association Rule Mining}
TS-ARM is a new paradigm, which treats a transaction database as a time series data. In line with this, the formal definition of the NARM problem needs to be redefined. In the TS-ARM, the association rule is defined as an implication:
\begin{equation}
    X(\Delta t)\implies Y(\Delta t),
\end{equation}
where $X(\Delta t)\subset O$, $Y(\Delta t)\subset O$, and $X(\Delta t)\cap Y(\Delta t)=\emptyset$. The variable $\Delta t=[t_1,t_2]$ determines the sequence of the transactions arisen within the interval $t_1$ and $t_2$, where $t_1$ denotes the start and $t_2$ the end time of the observation. The measures of support and confidence are redefined as follows:
\begin{equation}
\footnotesize
\mathit{conf_t}(X(\Delta t) \implies Y(\Delta t)) = \frac{n(X(\Delta t) \cap Y(\Delta t))}{n(X(\Delta t))}, 
\label{Eq:1}
\end{equation}
\begin{equation}
\footnotesize
\mathit{supp_t}(X(\Delta t) \implies Y(\Delta t)) = \frac{n(X(\Delta t) \cap Y(\Delta t))}{N(\Delta t)}, 
\label{Eq:2}
\end{equation}
where $\mathit{conf_t}(X(\Delta t) \implies Y(\Delta t))\geq C_{\max}$ and $\mathit{supp_t}(X(\Delta t) \implies Y(\Delta t))\geq S_{\max}$ denotes the confidence and support of the association rule $X(\Delta t)\implies Y(\Delta t)$ within the same time interval $\Delta t$. 

Let us highlight Eq.~(\ref{Eq:2}) with the following example: Let us assume the itemset is given as follows:
\begin{equation*}
\begin{aligned}
X([12,14])=&\{\mathit{MIN\_TEMPERATURE}\_18,\\
    & \mathit{MAX\_TEMPERATURE}\_20\},
\end{aligned}
\end{equation*}
and the transaction database captures features of passed 5 days, where each day is divided into 24 classes (i.e., total 120 transactions). If 2 matches in temperatures between $18^\circ C$ and are $20^\circ C$ are found in 5 days within the specified time interval $[12,14]$, the itemset has support $supp([12,14])=\frac{2}{5}=0.4$. 

The other aforementioned NARM measures (i.e., inclusion and amplitude) are independent on time and, consequently, they are employed in their original form.

\subsubsection{Representation of solutions}
The individuals in the nature-inspired algorithms $\mathbf{x}^{(t)}_i$ for $i=1,\ldots,Np$ are encoded as a real-valued vector (genotype):
\begin{equation}
\small
\begin{aligned}
    \mathbf{x}^{(g)}_i=&\{\langle\underbrace{x^{(g)}_{i,1},\ldots,x^{(g)}_{i,4}}_{Feat_{i,1}}\rangle,\ldots,\langle\underbrace{x^{(g)}_{i,13},\ldots,x^{(g)}_{i,16}}_{Feat_{i,4}}\rangle, \\
    & \langle\underbrace{x^{(g)}_{i,17},x^{(g)}_{i,18}}_{\Delta t_i}\rangle,\underbrace{x^{(g)}_{i,19}}_{Cp_i}\},
\end{aligned}
\end{equation}
where each element $x^{(g)}_{i,j}$ for $j=1,\ldots,16$ determines four quadruples determining the compound features $Feat^{(g)}_k$ for $k=1,\ldots,4$ into the transaction database, $\Delta t_i$ denotes the $i$-th time interval, $Cp_i$ the cutting point, and $g$ is the generation number. Thus, each numerical feature $\textit{Feat}^{(g)}_{\pi_j}$ consists of four real-valued elements decoded (phenotype) as:
\begin{equation}
\small
    \textit{Feat}^{(g)}_{\pi_j}=\left \{ \begin{array}{ll}
    x^{(g)}_{i,4j}\mapsto \pi^{(g)}_j, & \text{permutation}, \\
    x^{(g)}_{i,4j+1}\mapsto y^{(g)}_{\pi_j}, & \text{lower bound}, \\
    x^{(g)}_{i,4j+2}\mapsto z^{(g)}_{\pi_j}, & \text{upper bound}, \\
    x^{(g)}_{i,4j+3}\mapsto \mathit{Th}^{(g)}_{\pi_{4j}}, & \text{threshold value}, \\
    \end{array}\right .
\end{equation}
where permutation $\Pi=(\pi_1,\ldots,\pi_m)$ served for modifying the position of the feature within the association rules. Technically, all first elements denoting the corresponding features are sorted in descendent order, 
\begin{equation*}
     x^{(g)}_{i,4\pi_{1}}\leq x^{(g)}_{i,4\pi_2}\leq x^{(g)}_{i,4\pi_3}\leq x^{(t)}_{i,4\pi_4},
\end{equation*}
while their ordinal values determine their position in the permutation.

The two middle elements within quadruple encode a real-valued interval of feasible values $[lb^{(g)}_{\pi_j},ub^{(g)}_{\pi_j}]$ expressed as:
\begin{equation*}
\footnotesize
    lb^{(g)}_{\pi_j}=\left \{ \begin{array}{ll} 
    \left ( \textit{Ub}_{\pi_j}-\textit{Lb}_{\pi_j} \right ) x^{(g)}_{i,4\pi_j+1}, & \text{if}~x^{(g)}_{i,4\pi_j+1}<x^{(g)}_{i,4\pi_j+2}, \\
    \left ( \textit{Ub}_{\pi_j}-\textit{Lb}_{\pi_j} \right ) x^{(g)}_{i,4\pi_j+2}, & \text{otherwise},
    \end{array} \right .
\end{equation*}
and
\begin{equation*}
\footnotesize
    ub^{(g)}_{\pi_j}=\left \{ \begin{array}{ll} 
    \left ( \textit{Ub}_{\pi_j}-\textit{Lb}_{\pi_j} \right ) x^{(g)}_{i,4\pi_j+2}, & \text{if}~x^{(g)}_{i,4\pi_j+1}<x^{(g)}_{i,4\pi_j+2}, \\
    \left ( \textit{Ub}_{\pi_j}-\textit{Lb}_{\pi_j} \right ) x^{(g)}_{i,4\pi_j+1}, & \text{otherwise},
    \end{array} \right .
\end{equation*}
where $\textit{Lb}_{\pi_j}$ and $\textit{Ub}_{\pi_j}$ denote the lower and the upper values of the particular feature as found in the transaction database. 

The threshold value denotes the presence or absence of the feature $\textit{Feat}^{(g)}_{\pi_j}$ in the observed association rule according to the following equation:
\begin{equation*}
    \mathit{Th}^{(g)}_{\pi_j}=\left \{ \begin{array}{lc}
    \mathit{enabled}, & \text{if}~\textit{rand}(0,1)<  x^{(\mathit{g})}_{i,4\pi_j+2} \\ 
    \mathit{disabled}, & \text{otherwise}, \\
    \end{array}\right .
\end{equation*}
where $\textit{rand}(0,1)$ draws a value from uniform distribution in interval $[0,1]$.

The time interval $\Delta t$ is calculated according to the following expression:
\begin{equation}
\footnotesize
\Delta t= \begin{cases}
    \begin{array}{ll}
        \left[\lfloor K\cdot x^{(g)}_{17}\rfloor,\lfloor K\cdot x^{(g)}_{18}\rfloor\right], & \text{if}~x^{(g)}_{i,17}<x^{(g)}_{i,18}, \\
        \left[\lfloor K\cdot x^{(g)}_{18}\rfloor,\lfloor K\cdot x^{(g)}_{17}\rfloor\right], & \text{otherwise},
    \end{array}
\end{cases}
\end{equation}
where $K$ denotes the number of classes.

As the last element, the so-called cutting point is added to each vector that distinguishes the antecedent of the rule from the consequent ones. The cutting point $Cp$ is expressed as:
\begin{equation}
    \mathit{Cp}_i=\lfloor x^{(g)}_{i,19}\cdot (4-1)\rfloor +1,
\end{equation}
where $\mathit{Cp}_i\in[1,3]$. 

Finally, the results of this so-called genotype-phenotype mapping, where the values encoded into genotype are decoded into phenotype, is association rule $X\implies Y$ consisting of antecedent $X$ and consequent $Y$ separated by an implication sign positioned at the point determined by the variable $Cp$.

\subsubsection{Definition of the fitness function}
We tailored the fitness function presented in~\cite{fister2018differential} to deal with time series data as follows:
\begin{equation}
\footnotesize
\begin{aligned}
    f(\mathbf{x}^{(t)}_i)=&\frac{\alpha\cdot\mathit{supp}(X\Rightarrow Y^{(t)}_i)+\beta\cdot\mathit{conf}(X\Rightarrow Y^{(t)}_i)}{\alpha+\beta+\gamma+\delta}+\\
    & \frac{\gamma\cdot\mathit{incl}(X\Rightarrow Y^{(t)}_i)+\delta\cdot\mathit{ampl}(X\Rightarrow Y)}{\alpha+\beta+\gamma+\delta},
\end{aligned}
    \label{eq:fit}
\end{equation}
where $\alpha$, $\beta$, $\gamma$, and $\delta$ denote weights of the support, the confidence, the inclusion, and the amplitude of the association rule $X\Rightarrow Y$ decoded from the vector $\mathbf{x}^{(t)}_i$.

\section{Results}\label{sec:4}
The goal of the experimental study was two-fold: (1) to analyse a behavior of the system in smart agriculture, and (2) to show that the nature-inspired algorithms for TS-NARM can be applied in smart agriculture. In line with this, an experimental environment was established as illustrated in the last section, which enable creating a transaction database. Then, the nature-inspired algorithms for TS-NARM were applied to searching for hidden relationships between features in the transaction database. 

Two experiments were conducted in order to justify our hypotheses:
\begin{itemize}
    \item analysis of a behavior of the system in smart agriculture, 
    \item comparative study of five nature-inspired algorithms for TS-ARM.
\end{itemize}
In the remainder of the paper, the experimental setup is reviewed, then the algorithm configurations are discussed, and finally, the results of the aforementioned experiments are illustrated. 

\subsection{Experimental setup}
\begin{center}
\begin{figure}[!ht]
\includegraphics[width=8cm]{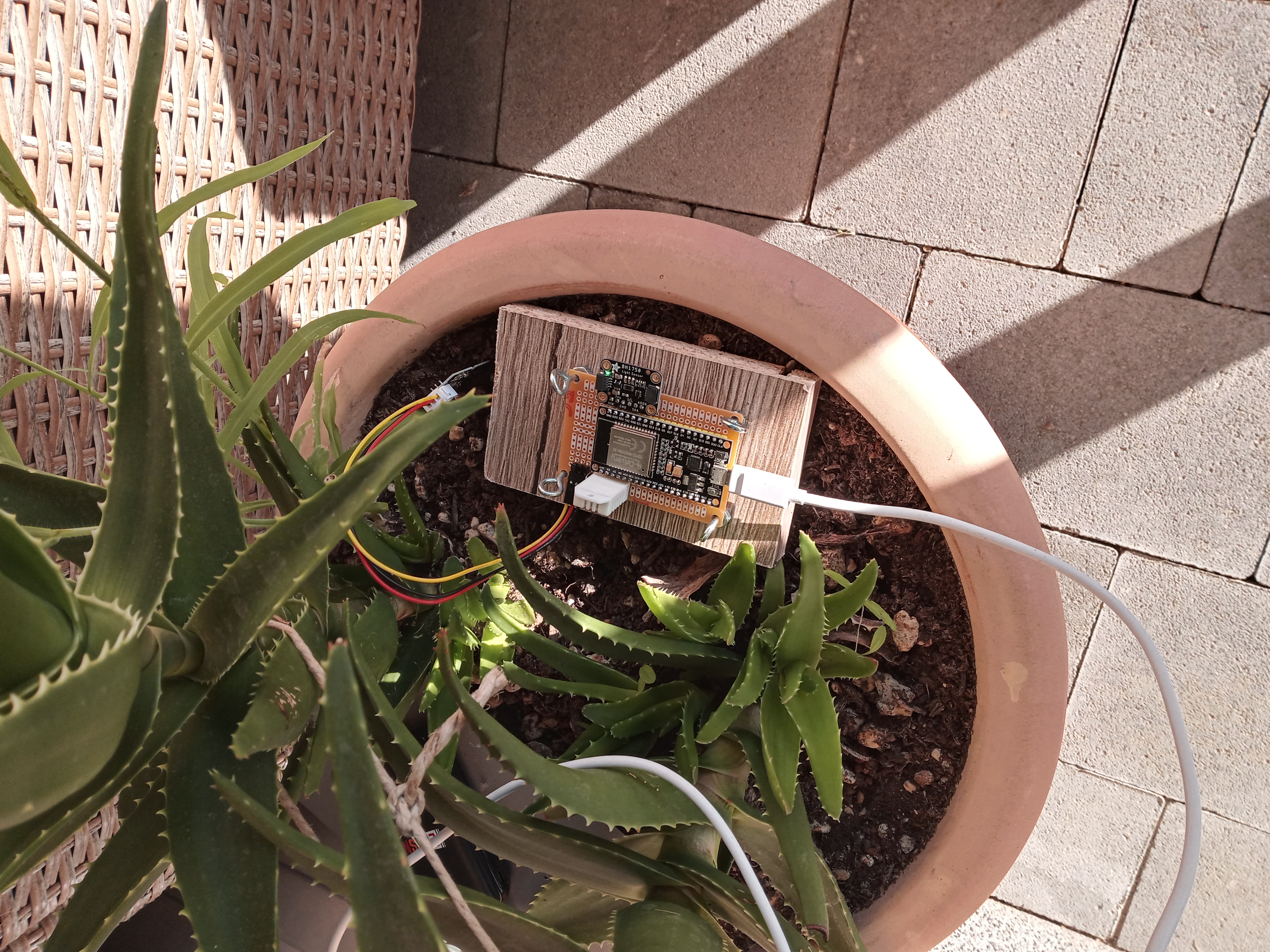}
\caption{Experimental environment}
\label{hardware}
\end{figure}
\end{center}

For the purpose of our study, Aloe Vera plant served as a plant for simulation of our smart agriculture concept. As can be seen at the Fig~\ref{hardware}, a rural network is built using sensors connected directly to the ESP32 NodeMCU control process unit. The unit is powered by a power bank of 20000 mAh capacity.
Three sensors for light, air temperature and humidity, and moisture sense land characteristics and transmit sensor data in approximately 5 sec intervals. The sensor data form time series of duration 1 hour. This means, that each time frame (also transactions) bears characteristics of $12~/\min\times 60~\min=720$ sensor data, in other words $T=720$. 

In summary, the transaction database contains data accumulated in 14 days. Consequently, it consists of $14\times 24=336$ different transactions.

\subsection{Algorithm configurations}
In our study, five nature-inspired algorithms were applied as follows: DE, GA, PSO, LSHADE and jDE. Thus, all implementations of algorithms were taken from the NiaPy library, where default parameters were taken from NiaPy examples~\cite{vrbancic2018niapy} (Table~\ref{tab:param}). 
\begin{table*}[htb]
\centering
    \caption {Parameter setting of the nature-inspired algorithms.}
    \label{tab:param}
    \begin{tabular}{ll}
    \hline
    Algorithm & Parameter setting \\ \hline
    DE & $F=0.5$. $CR=0.9$ \\ 
    GA & $p_m=0.01$, $p_c=0.8$ \\ 
    PSO  & $c_1=0.1$, $c_2=0.1$, $w=0.8$ \\ 
    LSHADE & $NP_{max}=18\cdot NP$, $NP_{min}=4\cdot NP$, $H=5$, $p=0.1$, $r^{arc}=2$ \\ 
    jDE & $F^{(0)}=0.5$, $CR^{(0)}=0.9$, $\tau=0.1$ \\ 
    \hline
    \end{tabular}
\end{table*}
The number of function evaluations for all algorithms was set to $MaxFEs=10,000$ and all algorithms had the population size of 50. We performed ten independent runs for each algorithm in test.

\subsection{Analysis of a behavior of the system in smart agriculture}
The system presents a cost-effective solution in smart agriculture that supports: data acquiring, data collection, and data preprocessing. Therefore, the purpose of the test was to analyse how the system behaves in the sense of the following system's quality metrics:
\begin{itemize}
    \item reliability,
    \item robustness,
    \item accuracy,
    \item scalability.
\end{itemize}
Indeed, the test comprises of evaluating three system components: hardware unit (data acquiring), data collection, and preprocessing. In line with this, the system underwent to continuous operating in duration of 14~days (Table~\ref{tab:stat}). 
\begin{table*}[htb]
\centering
    \caption {Data collection statistics.}
    \label{tab:stat}
    \begin{tabular}{ll}
    \hline
    Attribute & Value \\\hline
    Number of collected records transmitted onto the web & 233,980 \\ 
    Start time of collecting data & 2022-09-15, 00:00:04 \\ 
    End time of collecting data  & 2022-09-28,23:59:57 \\ 
    Collecting time period & approximately every 5 seconds \\ 
    Average collected records per day & 16,712 \\ 
    \hline
    \end{tabular}
\end{table*}
Thus, the acquired data from sensors are collected approximately each 5~seconds. In total, the system transmitted 233,980 records onto the web.

The results of data collection are depicted in Table~\ref{tab:sense},
\begin{table*}[htb]
\centering
    \caption {Time series data.}
    \label{tab:sense}
    \begin{tabular}{crrrrrr}
    \hline
MP & Light & Temperature & Humidity & Moisture & Date & Time \\ \hline
n1 & 0 & 24.70 & 57.90 & 1995 & 2022-09-15 & 00:00:04\\
n1 & 0 & 24.70 & 58.00 & 1991 & 2022-09-15 & 00:00:09\\
n1 & 0 & 24.70 & 58.20 & 1994 & 2022-09-15 & 00:00:14\\
n1 & 0 & 24.60 & 58.00 & 1993 & 2022-09-15 & 00:00:19\\
n1 & 0 & 24.60 & 58.00 & 1986 & 2022-09-15 & 00:00:25\\
n1 & 0 & 24.60 & 58.00 & 1991 & 2022-09-15 & 00:00:30\\
n1 & 0 & 24.60 & 58.00 & 1995 & 2022-09-15 & 00:00:35\\
n1 & 0 & 24.60 & 58.20 & 1993 & 2022-09-15 & 00:00:40\\
    \hline
    \end{tabular}
\end{table*}
from which it can be seen time series consisting of eight sensor data records acquired in 15.9.2022 starting at 00:00:04~AM. Each record consists of indicators obtained by light, temperature, humidity, and moisture sensors. The BH1750 light sensor provides 16-bit light measurements in lux, and measures light from 0 (night) to 100K lux (day). Temperature sensor senses temperature in range $-40^\circ C$ to $80^\circ C$. Humidity measuring range is in interval $0~\%$~RH to $100~\%$~RH with measurement accuracy of $\pm 2~\%$~RH. Soil moisture is detected by a simple water sensor, while the moisture values ranging from 0 to 2300. Data and time values are added by the web server.

As can be seen from Table~\ref{tab:sense}, all data were obtained from measuring point number~1 during the night due to value 0 measured by light sensor. The values from other sensors remained almost constantly, while the variances of their values could be ascribed to the measurement accuracy of the particular sensor.

Due to the big number of features obtained as a result of preprocessing, the illustration of the transactions saved into transaction database is omitted in the paper. Instead of this, the statistics of the preprocessed transactions is summarized in Table~\ref{tab:stat2}, 
\begin{table}[htb]
    \caption {Data preprocessing statistics.}
    \label{tab:stat2}
    \begin{tabular}{ll}
    \hline
    Total records in transaction database & 336 \\
    Total number of features & 18 \\
    Type of features & numeric \\
    \hline
    \end{tabular}
\end{table}
from which it can be seen that 336 transactions (time frames) emerged as a result of preprocessing.

\subsection{Comparative study}
The experiments was focused on evaluating the proposed nature-inspired algorithms for TS-ARM according to the standard ARM measures. The algorithms in the comparative study used parameter settings as illustrated in Table~\ref{tab:param}. The results of the experiments are illustrated in Table~\ref{tab:rez}
\begin{table*}[htb]
    \centering
    \caption{Rules found by the different algorithms.}
    \label{tab:rez}
    \begin{tabular}{l*{11}{r}}
        \toprule
        \multirow{2}{*}{Algorithm}  & \multicolumn{4}{c}  {Measures} & \multicolumn{2}{c}  {Lengths} & \multirow{2}{*} {Numrules} & \multirow{2}{*}{Intervals}  \\
                   &  supp  &  conf  & incl & ampl & antlen & conlen \\
        \midrule
         DE & \textbf{0.69} & \textbf{0.87} & 0.20 & 0.54 & 1.72 & 1.51 & 2,707 & 100~\% \\
         GA & 0.19 & 0.63 & \textbf{0.30} & 0.53 & \textbf{2.55}  & \textbf{2.30} & 40 & 96~\%\\
         PSO & 0.64 & 0.82 & 0.16 & \textbf{0.77} & 1.28 & 1.24 & \textbf{3,386} & 100~\% \\
         LSHADE & 0.59 & 0.84 & 0.16 & 0.74 & 1.27 & 1.24 & 2,588 & 96~\% \\
         jDE & 0.57 & 0.85 & 0.24 &0.37 & 2.09 & 1.73 & 664 & 100~\% \\
        \bottomrule  
    \end{tabular}
\end{table*}
depicting the achieved values according to four measures (i.e., support, confidence, inclusion, and amplitude), and average lengths of corresponding antecedent and consequent per each observed algorithm. Columns 'Numrules' and 'Intervals' are added to the table and denote the number of mined rules and the percentage of intervals covered by the rule, respectively. 

Interestingly, the best results according to support and confidence are distinguished by the DE, while the best results according to inclusion are achieved by the GA, and according to amplitude by the PSO. The longer length of features in antecedent and consequent are mined by the GA, where the length of both measures overcome the value of 2.30. The maximum number of rules were mined by the PSO (i.e., $3,386$), while the minimum by the GA (only $40$). As a matter of fact, all algorithms excellent cover the intervals in the rules.
    
\section{Discussion, Conclusions and further research}\label{sec:5}
The following conclusions can be obtained, according to the results of the first test: In general, the conducted test showed that the system is reliable due to the continuous operating over 14~days. During this period, it underwent different conditions (e.g., stormy, rain, sunny, etc.), and more day-night cycles. This fact justify that the system is also robust. Although the applied sensor are low-cost, the acquired data are accurate, especially, by considering the fact that errors can be compensated by averaging values of the big number of measurements. Finally, the system is scalable, because more sensors can be connected to the hardware unit and thus improve capturing of the land conditions.

The following conclusions may summarize the results of the second test carried out: The DE is excellent in searching for rules, where there exist good relationships between features regarding either other feature or the total number of transactions, respectively. The best use of the number of features in antecedent and consequent is identified by the GA, while the best covering of the numeric intervals is achieved by the PSO. On the other hand, the GA discovered the less number of association rules comparing with the other algorithm in test. Indeed, the highest number of rules is mined by the PSO. Consequently, the higher the number of mined rules, the better support and confidence, and contrary, the smaller the number of mined rules, the richer the association rules in the sense of the number of features in antecedent and consequent.

However, there are also several bottlenecks that were found when running experiments. All blockages are summarized as follows:
\begin{itemize}
\item Some intervals are occasionally omitted, and after the run, there are no rules linked to a specific interval.
\item Sometimes algorithms identify a rule with very high fitness, consequently, the algorithm falls within the local optimum, and after that, it is tough to find good rules in the other intervals.
\item After the initial experiments, we found that it is essentially to ensure more evaluations since they ensure that we find rules in different intervals.
\end{itemize}
In the future, it would be necessary to find a better local search or switch between different intervals to capture as much association rules as possible. It is recommended that a new metric being added to the fitness function, which would also control how much of the intervals are covered in the final results.

\backmatter

\section*{Declarations}

\bmhead{Funding} This work was supported by the Slovenian Research Agency (Research Core Funding Nos. P2-0057, P5-0027). This work has also been partially supportted through project PID2020-115454GB-C21 of the Spanish Ministry of Science and Innovation (MICINN).

\bmhead{Code and data availability}
The datasets and source codes are available from the corresponding author on reasonable request.

\bmhead{Conflict of interest} The authors declare that they have no potential conflict of interest.

\bibliographystyle{plainnat}
\bibliography{bibtex}



\end{document}